\documentclass[runningheads]{llncs}
\usepackage[T1]{fontenc}
\usepackage{graphicx}

\usepackage{color}
\usepackage{hyperref}
\usepackage{booktabs}
\usepackage{multirow}
\usepackage[inline]{enumitem}
\usepackage{amsmath}
\usepackage{dcolumn}
\usepackage{pifont}
\usepackage{pgfplots}
\usepackage{tikz}
\usepackage{CJK}
\usepackage{arydshln}

\definecolor{frenchblue}{rgb}{0.0, 0.45, 0.73}
\definecolor{babyblue}{rgb}{0.54, 0.81, 0.94}
\definecolor{classicrose}{rgb}{0.98, 0.8, 0.91}
\definecolor{beige}{rgb}{0.96, 0.96, 0.86}
\definecolor{forestgreen}{HTML}{2e7d43}

\definecolor{blue1}{HTML}{91BBE6}
\definecolor{blue2}{HTML}{3F90E0}
\definecolor{blue3}{HTML}{316FAD}

\definecolor{color1}{HTML}{FF9999}
\definecolor{color2}{HTML}{FF6666}
\definecolor{color3}{HTML}{FF3333}
\definecolor{color4}{HTML}{E60000}
\definecolor{color5}{HTML}{B30000}
\definecolor{color6}{HTML}{8CD98C}
\definecolor{color7}{HTML}{53c653}
\definecolor{color8}{HTML}{39ac39}
\definecolor{color9}{HTML}{2d862d}
\definecolor{color10}{HTML}{206020}
\definecolor{color11}{HTML}{cca300}

\begin{document}
\usepgfplotslibrary{groupplots}
\title{RIGHT: Retrieval-augmented Generation for Mainstream Hashtag Recommendation}
\titlerunning{RIGHT}
%
%
\author{Run-Ze Fan\inst{1,2}\orcidID{0000-0002-8505-7756} \and
Yixing Fan\inst{1,2}\orcidID{0000-0003-4317-2702} \and \\
Jiangui Chen\inst{1,2}\orcidID{0000-0002-6235-6526} \and
Jiafeng Guo\inst{1,2}\thanks{Corresponding author}\orcidID{0000-0002-9509-8674} \and \\
Ruqing Zhang\inst{1,2}\orcidID{0000-0003-4294-2541} \and
Xueqi Cheng\inst{1,2}\orcidID{0000-0002-5201-8195}
}

%
\authorrunning{R. Fan et al.}
%
\institute{CAS Key Lab of Network Data Science and Technology, ICT, CAS \and
University of Chinese Academy of Sciences, Beijing, China \\
\email{fanrunze21s@ict.ac.cn}\\
\email{\{fanyixing, chenjiangui18z, guojiafeng, zhangruqing, cxq\}@ict.ac.cn}
}

%
\maketitle              
%
\begin{abstract}
Automatic mainstream hashtag recommendation aims to accurately provide users with concise and popular topical hashtags before publication. 
Generally, mainstream hashtag recommendation faces challenges in the comprehensive difficulty of newly posted tweets in response to new topics, and the accurate identification of mainstream hashtags beyond semantic correctness.
However, previous retrieval-based methods based on a fixed predefined mainstream hashtag list excel in producing mainstream hashtags, but fail to understand the constant flow of up-to-date information. 
Conversely, generation-based methods demonstrate a superior ability to comprehend newly posted tweets, but their capacity is constrained to identifying mainstream hashtags without additional features.
Inspired by the recent success of the retrieval-augmented technique, in this work, we attempt to adopt this framework to combine the advantages of both approaches.
Meantime, with the help of the generator component, we could rethink how to further improve the quality of the retriever component at a low cost.
Therefore, we propose \textit{\textbf{R}etr\textbf{I}eval-augmented \textbf{G}enerative Mainstream \textbf{H}ash\textbf{T}ag Recommender} (\textbf{RIGHT}), which consists of three components:
\begin{enumerate*}[label=(\roman*)]
    \item a retriever seeks relevant hashtags from the entire tweet-hashtags set;
    \item a selector enhances mainstream identification by introducing global signals;
    and \item a generator incorporates input tweets and selected hashtags to directly generate the desired hashtags.
\end{enumerate*}
The experimental results show that our method achieves significant improvements over state-of-the-art baselines. 
Moreover, RIGHT can be easily integrated into large language models, improving the performance of ChatGPT by more than 10\%.
Code will be released at: \url{https://github.com/ict-bigdatalab/RIGHT}.

\end{abstract}
\section{Introduction}
\label{sec:introduction}

Millions of user-generated microblogs flood Twitter daily, surpassing users' comprehension. 
To facilitate rapid and easy understanding, hashtags (e.g., \textit{\#ChatGPT}) are extensively used to convey central ideas and topics, which also enhance content visibility to reach a broader audience~\cite{kwak2010twitter}. 
Such hashtags are commonly referred to as mainstream hashtags, denoting their status as not only the most prevalent hashtags but also semantically accurate.
For instance, in the context of Kobe Bryant's untimely demise, both \textit{\#KobeDead} and \textit{\#KobeDeath} can be utilized, with both possessing accurate semantic meanings.
However, the former holds more widespread usage and has achieved the distinction of being a mainstream hashtag.

\begin{figure}[t]
\centering
\makeatother\def\@captype{figure}\makeatother
	\centering
    \vspace{-10pt}
	\includegraphics[width=1\columnwidth]{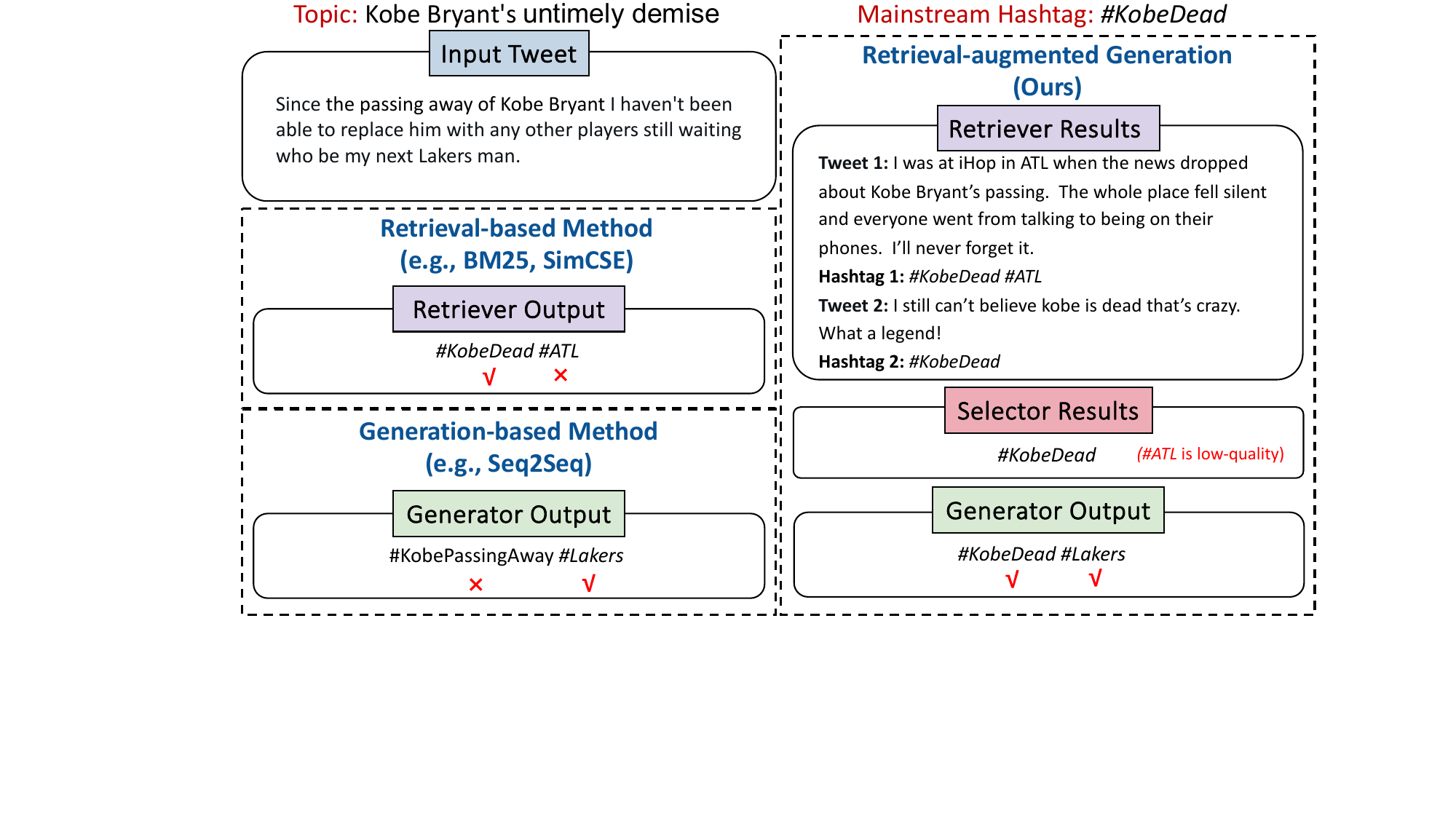}
 \vspace{-15pt}
    \caption{Illustration of evaluating hashtag recommendation with different methods.}
    \label{fig:main}
\vspace{-15pt}
\end{figure}
To provide mainstream hashtags, two main challenges need to be addressed.
First, comprehending a new tweet presents challenges primarily attributable to the absence of real-time information~\cite{wang-2019-microblog,DBLP:journals/corr/abs-2104-08723}. This is a direct consequence of the continuous emergence of numerous new tweets in response to new topics and events.
Second, accurately identifying mainstream hashtags beyond semantic correctness remains a challenging task. The reason is that numerous hashtags could be used to describe a topic, but only a few are mainstream.

To address the above challenges, a considerable amount of work has been proposed, which could be divided into two research lines~\cite{ding-2012-automatic,gong-2015-hashtag,weston-2014-tagspace,DBLP:conf/ijcai/GongZ16,wang-2019-microblog}.
Retrieval-based methods retrieve hashtags from a fixed predefined mainstream hashtag list~\cite{weston-2014-tagspace,DBLP:conf/ijcai/GongZ16}, which could alleviate the second problem.
However, their ability to fully grasp the meaning of a newly posted tweet in response to emerging topics and events is constrained. Moreover, it is a considerable cost to maintain the predefined list~\cite{wang-2019-microblog}.
In contrast, generation-based methods~\cite{wang-2019-microblog,DBLP:journals/corr/abs-2104-08723,MAO2022109581} demonstrate remarkable proficiency in comprehending new tweets and generating semantically accurate hashtags, owing to their substantial pretraining knowledge.
Nevertheless, they may encounter difficulties when it comes to identifying mainstream hashtags without enough mainstream information.
As a result, the tweet might fail to be indexed by a mainstream hashtag on microblog services due to the tags' unpopularity, weakening the recall rate of microblog searches. 
Inspired by the recent success of retrieval-augmented generation technique~\cite{asai-etal-2023-retrieval,ramos-2023-retrieval,kim-2021-structure,gao-2022-retrieval,zhang-2022-domain}, therefore, we try to adapt this method to mainstream hashtags recommendation, utilizing the advantages of both retrieval and generation approaches.

Typically, retrieval-augmented techniques incorporate the results of the retriever, whether explicitly or implicitly, into the generator to enhance the quality of generation.
Utilizing this framework, the introduction of a generator endowed with strong comprehensive capabilities might mitigate the dependency on the quality of the retriever~\cite{mialon2023augmented}. 
Thus, we could rethink the trade-off between the quality and the cost of the retriever. 
Traditional retrieval-based methods rely on a predefined list of mainstream hashtags, which can ensure the quality of the retrieved information, but maintains such a list at a significant cost.
To reduce the maintenance burden, we transform the small predefined list into a larger aggregation of existing tweet-hashtags pairs, which can be automatically collected and updated without manual cost.
However, this approach carries the risk of introducing numerous low-quality hashtags due to the informal characteristics of social media content.
Such hashtags have the potential to mislead the generator.
As illustrated in Figure~\ref{fig:main}, both \textit{\#KobeDead} and \textit{\#ATL} are results of the retriever.
Nonetheless, it is noteworthy that \textit{\#ATL} is law-quality, even though tweet $1$ exhibits the highest degree of similarity with the input tweet. 
Consequently, it becomes imperative to further improve the quality of retrieved information without increasing the cost.

Therefore, in this study, we propose a \textit{\textbf{R}etr\textbf{I}eval-augmented \textbf{G}enerative Mainstream \textbf{H}ash\textbf{T}ag Recommender}~(\textbf{RIGHT}), which combines the retriever and the generator by the retrieval-augmented technique with inserting a selector.
Specifically, our method involves three components: 1) \textbf{Retriever} is utilized to acquire relevant hashtags. We retrieve the tweets most similar to the input from the tweet-hashtags corpus and obtain the corresponding hashtags set.
2) \textbf{Selector} is used to improve the capability of identifying mainstream hashtags.
we incorporate three features, the similarity between the input tweet and the retrieved tweet and its hashtags, as well as the frequency of the hashtags, to enhance the mainstream information.
3) \textbf{Generator} is leveraged to provide strong semantic comprehension and the ability of hashtag generation.
We concatenate the selected hashtags with the input tweet and feed it into the generator to obtain the desired hashtags.
In this way, we can utilize not only the retriever and the selector to seek the mainstream hashtags but also the generator to produce the desired hashtags flexibly.

We conduct experiments on two large-scale datasets (i.e., English Twitter (THG) and Chinese Weibo (WHG)). 
Experimental results show that our method achieves significant improvements over state-of-the-art baselines. 
Moreover, as it can be easily incorporated in black-box language models, we also apply our framework to ChatGPT by zero-shot instruction learning, bringing a 12.7\% boost for THG and 18.3\% for WHG in F1@1. 
Finally, to deeply understand this method, we present a detailed analysis.

\section{Methodology}
\label{sec:methodology}

We propose \textit{\textbf{R}etr\textbf{I}eval-augmented \textbf{G}enerative Mainstream \textbf{H}ash\textbf{T}ag Recommender} (\textbf{RIGHT}), a simple yet effective framework for mainstream hashtag recommendation, which includes three components: retriever, selector, and generator.

\begin{figure}[t]
\centering
\makeatother\def\@captype{figure}\makeatother
	\centering
    \vspace{-10pt}
	\includegraphics[width=1\columnwidth]{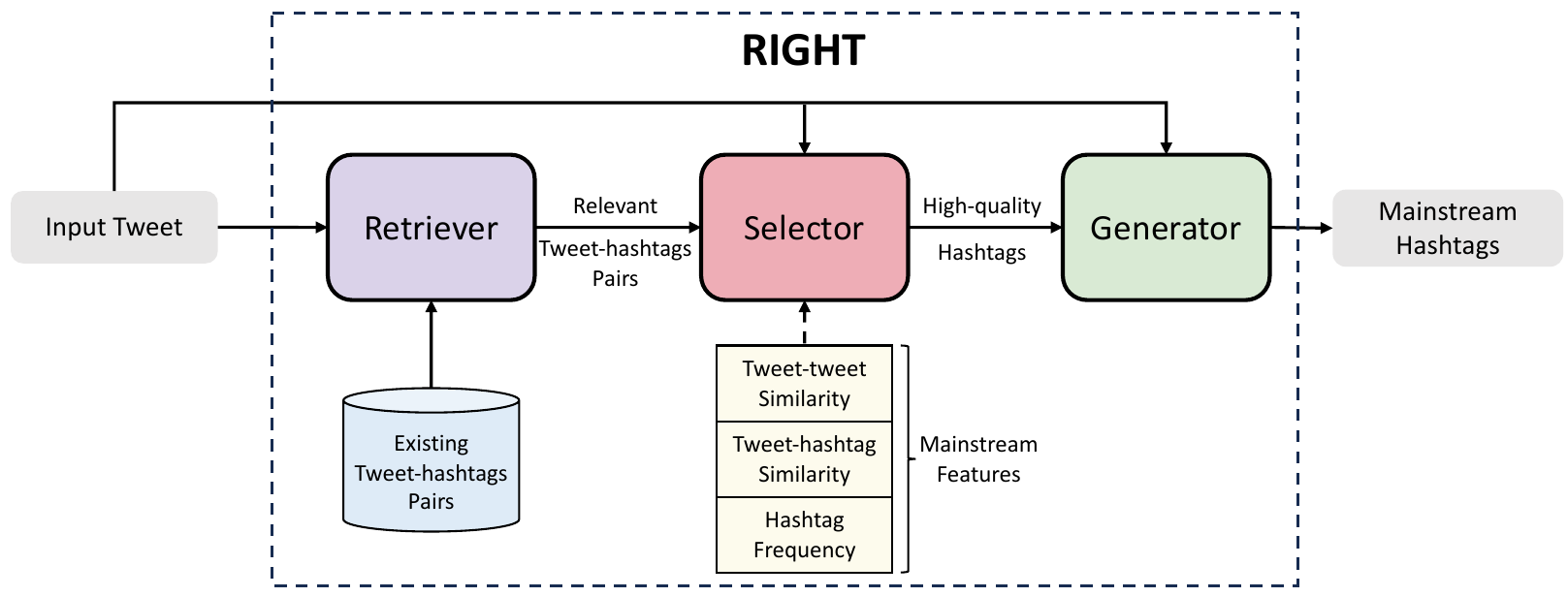}
    \caption{Our RIGHT framework consists of a retriever, a selector, and a generator.}
    \label{fig:framework}
\vspace{-15pt}
\end{figure}

Overall, we first utilize the retriever to retrieve the tweets most similar to the input from the existing tweet-hashtags corpus and obtain the corresponding hashtags set. Then, we adopt a selector to select the hashtags that are most probable mainstream from the retrieved labels using three signals. Finally, We concatenate the selected hashtags with the input tweet and feed it into the generator to obtain the desired hashtags. An overview of our method is shown in Figure~\ref{fig:framework}.

\subsection{Retriever}
\label{sec:retriever}
The goal of the retriever is to retrieve the top-$N$ tweet-hashtags pairs on the same topic with the input tweet from the existing corpus, aiming to find relevant hashtags on the same topic.

Inspired by Wang et al.~\cite{wang-2022-training}, we view the labeled training data as our corpus and index these as input-label pairs, i.e., $\mathcal{C} = \{ (\tilde{t}_i, \tilde{H}_i) \}$. Then, given the input tweet $t$, the retrieval model $\mathcal{R}$ matches it with all tweets in the corpus and returns the top-$N$ most similar tweet-hashtags pairs together with their scores:
\begin{equation}
    \{(\tilde{t}_1, \tilde{H}_1, \tilde{s}_1), \dots, (\tilde{t}_N, \tilde{H}_N, \tilde{s}_N)\} = \mathcal{R}(t|\mathcal{C}), \nonumber
\end{equation}
where we denote $\tilde{s}_i$ as the similarity between $t$ and the $i$-th retrieved tweet $\tilde{t}_i$. Each $\tilde{H}_i$ consists of hashtags $\{{\tilde{h}_1^i}, \dots, {\tilde{h}_{\vert \tilde{H}_{i}\vert}^{i}}\}$. We report the results with sparse retrieval (e.g., BM25~\cite{DBLP:conf/sigir/RobertsonW94}) and dense retrieval (e.g., SimCSE~\cite{gao-2021-simcse}) in experiments.

\subsection{Selector}
\label{sec:selector}
The goal of the selector is to filter the low-quality and non-mainstream hashtags existing in the results of the retriever (see Figure~\ref{fig:main}). We consider three mainstream features: the similarity between the input tweet and the retrieved tweet and its hashtags, as well as the frequency of the hashtags.
Thus, we train a selector to compute the similarity between the tweet and the hashtags and propose a simple algorithm for hashtag ranking.

\subsubsection{Training.}
The training data consists of positive samples and hard negative samples. Each hashtag labeled in a tweet can be viewed as a positive sample $t^+$.
However, a significant challenge lies in constructing hard negative samples ($t^-$) to facilitate the efficient selection of mainstream hashtags on the same topic by the selector. Inspired by BERT~\cite{devlin-2019-bert}, we propose to create a hard negative sample by disturbing the labeled hashtag without changing the semantic meaning. Specifically, we randomly select a word to:
\begin{enumerate*}[label=(\roman*)]
    \item replace with its synonym 70\% of the time;
    \item delete 10\% of the time;
    \item swap with the adjacent word 10\% of the time;
    \item insert a synonym after it 10\% of the time.
\end{enumerate*}
Thus, we obtain a training dataset $\{ (t_i, t_i^+, t_i^-)|i=1,\dots,N\}$.
Finally, we utilize contrastive learning to train our selector by minimizing the following loss:
\begin{equation}
    \mathcal{L}_{\mathcal{S}} = -\log \frac{e^{\mathrm{sim}(\mathbf{h}_{t_i}, \mathbf{h}_{t_i}^+) / \tau}}
        {\sum_{j=1}^{L}{\left( e^{\mathrm{sim}(\mathbf{h}_{t_i}, \mathbf{h}_{t_j}^+) / \tau} + e^{\mathrm{sim}(\mathbf{h}_{t_i}, \mathbf{h}_{t_j}^-) / \tau}\right) }}\nonumber
\end{equation}
where $\mathrm{sim}$ presents similarity, $\mathbf{h}_{t}$ indicates the representation of $t$, $L$ is mini-batch size, and $\tau$ is a temperature hyperparameter.

\subsubsection{Inference.}
In the inference stage, we propose a simple algorithm for hashtag ranking. Given an input tweet $t$ and the result of the retriever $\{(\tilde{t}_1, \tilde{H}_1, \tilde{s}_1), \dots, \\(\tilde{t}_N, \tilde{H}_N, \tilde{s}_N)\}$, we put retrieved hashtags into a set $\{\tilde{h}_1, \dots, \tilde{h}_M\}$ where we denote $M$ as the number of different hashtags, and record the number of occurrences $\{ f_1, \dots, f_M \}$ and the corresponding score of each retrieved hashtag $\{ \tilde{s}_{i,1} \dots, \tilde{s}_{i,f_i}  \}$. Then, we match the input tweet $t$ with all hashtags using the selector $\mathcal{S}$ to obtain the similarity score between the tweet and all hashtags $\{ \ddot{s}_1, \dots, \ddot{s}_M \}$:
\begin{equation}
    \ddot{s}_m = \mathcal{S}(t, \tilde{h}_m).
\end{equation}
Finally, we average the tweet-to-tweet similarity for each hashtag and add the similarity between the tweet and the hashtag. Since hashtags that occur more frequently are more likely to be mainstream, we magnify the sum of the similarity score and ranking score with a downscaled frequency:
\begin{equation}
    s_i = (({\frac{1}{f_i}}{\sum_{j=1}^{f_i}{\tilde{s}_{i,j}}}) + \ddot{s}_i) \times \left(1+  \left((f_i-1) / 10\right)\right),\nonumber
\end{equation}
and sort the hashtags by the final score from largest to smallest to select the top-$k$ hashtags $\{\tilde{h}_1, \dots, \tilde{h}_k\}$.

\subsection{Generator}
\label{sec:generator}
The goal of the generator is to generate the desired hashtags, given an input tweet $t$ and the selected hashtags $\{ \tilde{h}_1, \dots, \tilde{h}_k \}$. We concatenate the input tweet with the retrieved hashtags and separate each hashtag by a special token:
\begin{equation}
    I =<t, \mathtt{SEP1}, \tilde{h}_1, \mathtt{SEP1}, \tilde{h}_2, \dots, \mathtt{SEP1}, \tilde{h}_k >, \nonumber
\end{equation}
and feed it into the generator $\mathcal{G}$, which will output a concatenated sequence $O$ that includes the hashtags, with each hashtag separated by another token:
\begin{equation}
O=<h_1, \mathtt{SEP2}, h_2, \dots, \mathtt{SEP2} ,h_{\vert {H} \vert} >. \nonumber
\end{equation}
By easily splitting by the special token, we would obtain the hashtag list $H= \{ h_1, h_2,\dots, h_{\vert H \vert}\}$.

The generative model could be Transformer-based encoder-decoder architecture (e.g., T5~\cite{DBLP:journals/jmlr/RaffelSRLNMZLL20}, BART~\cite{DBLP:conf/acl/LewisLGGMLSZ20}) or decoder-only architecture (e.g., a series of GPT~\cite{radford2018improving,radford2019language,DBLP:conf/nips/BrownMRSKDNSSAA20}). Thus, the training stage focuses on the finetuning of generative models by minimizing the cross-entropy loss:
\begin{equation}
    \mathcal{L}_{\mathcal{G}} = \sum_{(I,O)\in \mathcal{D}}{-\log p(O|I;\theta_{\mathcal{G}})},\nonumber
\end{equation}
where $I$ is the input sequence consisting of the input tweet and the selected hashtags, $O$ is the output sequence consisting of the desired hashtags, and $\theta_{\mathcal{G}}$ is the parameters of the generator.

\begin{table*}[t]
 \centering
 \caption{Data statistics for the English Twitter hashtag generation (THG) dataset and the Chinese Weibo hashtag generation (WHG) dataset. \# T-H pairs denotes the number of tweet-hashtags pairs. \#AvgHashtags denotes the average number of hashtags in each tweet-hashtags pair. AvgTweetLen denotes the avenge length (token level) of all input tweets. AvgHashtagLen denotes the average length (token level) of all hashtags.}
 \vspace{-2pt}
 \resizebox{0.8\columnwidth}{!}{
 \begin{tabular}{lcccccc}\toprule
    \multirow{2}{*}{\bf Dataset} & \multicolumn{3}{c}{THG Dataset} & \multicolumn{3}{c}{WHG Dataset}
    \\ \cmidrule(lr){2-4}\cmidrule(lr){5-7}
             & Train  & Validation & Test    & Train  & Validation & Test \\\midrule
    \# T-H pairs    & 201444 & 11325 & 11328 & 307401 & 2000 & 2000\\
    \# AvgHashtags  & 4.1 & 4.1 & 4.1 & 1.0 & 1.0 & 1.0 \\
    AvgTweetLen & 39.7 & 39.6 & 39.6 & 87.1 & 86.8 & 87.8 \\
    AvgHashtagLen & 3.1 & 3.0 & 3.0 &  6.6 & 6.5 & 6.5 
    \\\bottomrule
 \end{tabular}
 }
 \vspace{-10pt}
 \label{tab:dataset}
\end{table*}
\section{Experiments}
\label{sec:experiment}

\subsection{Experimental Setup}
\subsubsection{Datasets.}
Our experiments are conducted on two large-scale datasets, which were crawled from official media and influencers of social media~\cite{MAO2022109581}. The details are shown in table~\ref{tab:dataset}.
\begin{itemize}
\item \textbf{THG:} The English Twitter hashtag generation (THG) dataset has been crawled from official Twitter sources, encompassing organizations, media outlets, and other authenticated users, with the primary objective of acquiring tweets of superior quality.
\item \textbf{WHG:} The Chinese Weibo hashtag generation (WHG) dataset has been acquired through the systematic extraction of microblogs from Weibo, encompassing notable sources including \textit{People’s Daily}, \textit{People.cn}, \textit{Economic Observe press}, \textit{Xinlang Sports}, and various other accounts boasting over 5 million followers. These accounts span diverse domains, encompassing politics, economics, military affairs, sports, and more.
\end{itemize}
We use the training datasets as our retrieval corpus.

\subsubsection{Evaluation Metric.}
Following previous work~\cite{wang-2019-microblog,MAO2022109581}, we utilize ROUGE metrics and F1 scores at $K$ as our evaluation metric. 
The average ROUGE score measures the overlap between the generated sequence of hashtags (excluding special tokens) and the reference sequence, including ROUGE-1, ROUGE-2, and ROUGE-L. 
For F1 scores at $K$, different $K$ values result in a similar trend, so only F1@1 and F1@5 are reported. 
We report results on the test dataset.
Noticeably, for the WHG dataset, where input posts have only one hashtag, F1@1 and F1@5 are identical, so we only report F1@1 for this dataset.

\subsubsection{Implementation Details.} Our implementation details of the retriever, selector, and generator are the following:
\begin{itemize}
    \item \textbf{For Retriever}, we utilize BM25~\cite{DBLP:conf/sigir/RobertsonW94} and SimCSE~\cite{gao-2021-simcse} (i.e., RoBERTa-Large~\cite{liu2019roberta} for THG and Bert-Base-Chinese~\cite{devlin-2019-bert} for WHG)
as our retrievers. Following Gao et al.~\cite{gao-2021-simcse}, we train our model for 3 epochs with a learning rate of 1e-5. The hyperparameter of $N$ is set to 10, and the batch size is 6 per device.
    \item \textbf{For Selector}, we use the training datasets from THG and WHG to construct our hard negative samples, which are subsequently employed for training our selectors in both English and Chinese independently. We utilize RoBERTa-Large for THG and Bert-Base-Chinese for WHG. The temperature $\tau$ is $0.05$ and other hyperparameters are the same as the retriever.
    \item \textbf{For Generator}, we fine-tune a T5-base~\cite{DBLP:journals/jmlr/RaffelSRLNMZLL20} for THG and a mT5-small~\cite{DBLP:conf/naacl/XueCRKASBR21} for WHG and use Adam~\cite{DBLP:journals/corr/KingmaB14} as an optimizer. We set the weight decay and batch size as 1e-5 and 16 and grid-search the learning rate, training epochs, and the number of concatenated hashtags $k$ from \{3e-4, 1e-4, 5e-5\}, \{5, 10\}, and \{1, 3, 5, 7, 9\} respectively. The maximum length is 180 for T5-base and 256 for mT5-small. The special token $\mathtt{SEP1}$ and $\mathtt{SEP2}$ are $\mathtt{<extra\_id\_0>}$ and $\mathtt{<extra\_id\_1>}$ respectively.
\end{itemize}
All models are trained on four NVIDIA Tesla K80.

\begin{table}[t]
\renewcommand{\arraystretch}{1.1}
 \centering
 \vspace{-2pt}
 \caption{The prompt used for ChatGPT. The Chinese version is its translation.}
 \vspace{-2pt}
 \begin{tabular}{l m{9cm}<{\centering}}  
 \toprule
    \textbf{Baseline} & \textbf{Instruction}
    \\ \midrule
    \textbf{ChatGPT} &  I want you to act as a hashtag annotator. I will provide you a tweet and your role is to annotate the relevant hashtag. You should use the related knowledge and find the topic. I want you only reply the hashtags segmented by ``\#'' and nothing else, do not write explanations. I want you segment the word in a hashtag by space. My first tweet is \{\texttt{Input Tweet}\}.\\ \hdashline
    $\mathrm{\mathbf{RIGHT}}^\mathrm{\mathbf{ChatGPT}}$ & I want you to act as a hashtag annotator. I will provide you with a tweet, and your role is to annotate the relevant hashtag. Using your related knowledge, you should identify the topic and reply with only the hashtags segmented by ``\#'', without any explanations.  Make sure to capitalize the first letter of the word. Make sure to split every word in a hashtag by a space. There are some potential hashtags:[\{\texttt{Retrieved Top-$k$ Hashtags}\}]. You can decide whether use the part of them or not. My first tweet is \{\texttt{Input Tweet}\}. 
    \\\bottomrule
 \end{tabular}
 \vspace{-8pt}
 \label{tab:instruction}
\end{table}

\subsubsection{Baselines.} Our baselines consist of retrieval-based methods, generation-based methods, and retrieval-augmented generative methods:
\begin{itemize}
\item \textbf{Retrieval-based methods:} Following Mao et al.~\cite{MAO2022109581}, we construct the predefined hashtags list from all hashtags in the training datasets and select top-4 hashtags for THG and top-1 for WHG according to the average number of hashtags in each data item. We apply BM25 and SimCSE to the hashtag recommendation: 
    \begin{enumerate*}[label=(\roman*)]
    \item \textbf{BM25}~\cite{DBLP:conf/sigir/RobertsonW94} is a traditional strong sparse retrieval based on term matching.
    \item \textbf{SimCSE}~\cite{gao-2021-simcse} is a representative dense retriever, which applies a simple contrastive learning framework to present sentence embeddings on semantic textual similarity tasks. We fine-tuned SimCSE by constructing positive samples and hard negative samples from BM25.
    \end{enumerate*}
\item \textbf{Generation-based methods:} We consider three predominant generative methods: 
\begin{enumerate*}[label=(\roman*)]
\item \textbf{ChatGPT}
is a powerful large language model to execute various NLP tasks~\cite{ko2023chatgpt}. Specifically, we adopt \texttt{gpt-3.5-turbo} and instruction zero-shot learning to evaluate our task (Prompts are shown in Table~\ref{tab:instruction}).
\item \textbf{SEGTRM Soft}~\cite{MAO2022109581} is the previous SOTA on our datasets, an end-to-end generative method segments selection-based deep transformer.
\item \textbf{Seq2Seq}~\cite{wang-2019-microblog} is the first generation-based method for hashtag recommendation. Due to the unreality to assume the existence of conversations before publishing the tweet~\cite{DBLP:journals/corr/abs-2104-08723} and the lack of conversation contexts, we reimplement the Seq2Seq model on the pretrained language model~(T5-base for THG and mT5-small for WHG) to formulate this task to a seq-to-seq paradigm.
\end{enumerate*}

\item \textbf{Retrieval-augmented Generative Methods (Ours):} We apply our\\ retrieval-augmented framework to ChatGPT by incorporating the retrieval results into the instruction to prompt the model to generate mainstream hashtags, denoted as $\mathrm{\mathbf{RIGHT}}^\mathrm{\mathbf{ChatGPT}}$ (Prompts are shown in Table~\ref{tab:instruction}). We only use the best retriever on the datasets (i.e., SimCSE for THG and BM25 for WHG), due to the high cost of ChatGPT. Moreover, we use BM25 and SimCSE as our retriever of RIGHT, denoted them as $\mathrm{\mathbf{RIGHT}}_\mathrm{\mathbf{BM25}}$ and $\mathrm{\mathbf{RIGHT}}_\mathrm{\mathbf{SimCSE}}$.

\end{itemize}

\begin{table*}[t]
 \small
 \centering
 \renewcommand{\arraystretch}{1}
  \vspace{-2pt}
 \caption{Main results (\%) on the THG and WHG datasets. Bold and \underline{underline} indicate the best and second method respectively. We donate ROUGE as RG. $*$ indicates statistically significant improvements over all baselines (p-value $< 0.05$).}
\vspace{-2pt}
 \resizebox{\columnwidth}{!}{
 \begin{tabular}{lccccccccc}\toprule
    \multirow{2}{*}{\bf Model} & \multicolumn{5}{c}{\textbf{THG}} & \multicolumn{4}{c}{\textbf{WHG}}
    \\ \cmidrule(lr){2-6}\cmidrule(lr){7-10}
             & RG-1 &  RG-2 & RG-L & F1@1 & F1@5    & RG-1 &  RG-2 & RG-L & F1@1  \\\midrule
    \multicolumn{10}{c}{\textit{Retrieval-based Methods}} \\ \midrule
    BM25 & 16.23 & 4.17 & 15.11 & 5.92 & 9.84 & 61.98 & 58.76 & 61.81 & 48.20\\ 
    SimCSE & 28.43 & 10.34 & 26.38 & 12.40 & 15.15 & 59.71 & 55.81 & 59.54 & 47.65 \\\midrule
    \multicolumn{10}{c}{\textit{Generation-based Methods}} \\ \midrule
    ChatGPT & 44.60 & 27.67 & 39.29 & 9.72 & 26.08 & 32.27 & 24.54 & 31.80 & 7.9 \\
    SEGTRM Soft  & 51.18 & 37.15 & 47.05 & 27.17 & 29.02 & 55.51 & 51.28 & 54.30 & 30.72\\
    Seq2Seq & 59.90 & 41.39 & 59.15 & 29.75 & 41.71 & 66.64 & 61.71 & 66.39 & 48.60 \\\midrule
    \multicolumn{10}{c}{\textit{Retrieval-augmented Generative Methods} (Ours)} \\ \midrule
    $\mathrm{RIGHT}^\mathrm{ChatGPT}$ & 47.54 & 25.63 & 44.47 & 22.39 & 31.09 & 48.17 & 41.51 & 47.75 & 26.15 \\
    $\mathrm{RIGHT}_\mathrm{BM25}$ & \underline{61.60} & \underline{43.77} & \underline{60.85} & \underline{30.27} & \underline{42.98} & \textbf{70.62}$^*$ & \textbf{66.12}$^*$ & \textbf{70.35}$^*$ & \textbf{53.85}$^*$\\
    $\mathrm{RIGHT}_\mathrm{SimCSE}$ & \textbf{62.11}$^*$ &\textbf{43.86}$^*$ & \textbf{61.39}$^*$ & \textbf{30.58}$^*$ & \textbf{43.23}$^*$ & \underline{68.84} & \underline{64.19} & \underline{68.56} & \underline{51.50}
    \\\bottomrule
 \end{tabular}
 }
 \vspace{-10pt}
 \label{tab:main_result}
\end{table*}

\subsection{Main Results}
\label{subsec:main_result}
As shown in Table~\ref{tab:main_result}, we can observe that:
\begin{enumerate}
    \item Among the retrieval-based methods, the performance of SimCSE outperforms BM25 in THG, while BM25 demonstrates superior performance in WHG. This difference may be attributed to that English hashtags tend to be concise summaries, while Chinese hashtags often comprise small sentences extracted directly from the input text. Consequently, dense retrieval approaches utilizing semantic matching may be more suitable for English datasets, while Chinese datasets may benefit more from sparse retrieval techniques based on term matching.
    \item Among the generation-based methods, Seq2Seq performs well on both datasets, potentially attributable to the utilization of mainstream hashtag knowledge from the training dataset during fine-tuning. However, ChatGPT lags behind other generation methods, suggesting a deficiency in mainstream hashtag knowledge despite its vast repository of general knowledge.
    \item Among the retrieval-augmented generative methods, retrieval augmentation brings the performance of baselines to a new level, demonstrating the effectiveness of our method. For ChatGPT, retrieval augmentation boosts F1@1 performance by 12.67\% for THG and 18.25\% for WHG, indicating the substantial value of mainstream hashtag knowledge. For RIGHT, both sparse and dense retrievers show the potential to enhance performance compared with Seq2Seq. Specifically, SimCSE is particularly effective for THG, while BM25 performs better for WHG. The reason could be attributed to the superiority in the performance of retrieval-based methods is directly proportional to the enhancement of the retrieval augmentation. Moreover, different retrievers excel in different scenarios, emphasizing the importance of the careful selection of retrievers based on specific use cases. The performance of the retrieval-based approach serves as a preliminary guide for informed decision-making.
\end{enumerate}

Overall, our method shows robustness across various scenarios, whether applied with a fine-tuned generation model or a large black box language model. Regardless of the retrieval approach used, our method consistently improves performance.

\subsection{Analysis}
\label{subsec:analysis}
\subsubsection{Ablation Study.}
We conduct an ablation study to explore the impact of each component in RIGHT on THG: 1) \textbf{w/o Retriever}: We remove the retriever and randomly concatenate $k$ hashtags from the training dataset with the input tweet. 2) \textbf{w/o Selector}: We remove the selector and directly use the top-$k$ hashtags from the retriever's results by the similarity between the input tweet and the retrieved tweet. 3) \textbf{w/o Generator}: We remove the generator and output the top-4 hashtags produced by the selector. Table~\ref{tab:ab_study} presents the results, indicating that:
\begin{enumerate}
\item The performance improvement is considerable through the integration of the retriever, confirming that the incorporation of mainstream hashtag knowledge indeed facilitates accurate hashtag selection.
\item Without the selector, the performance gains are limited, indicating that simply being on the same topic is insufficient. It is crucial to identify and incorporate mainstream hashtags.
\item The generator is crucial in RIGHT, emphasizing the significant impact of semantic comprehension on performance. In contrast to the retrieval-based approaches, it is more powerful to directly output the hashtags in the tweet-hashtags pair that are most similar to the input.
\end{enumerate}

\begin{table}[t]
 \centering
 \vspace{-2pt}
 \caption{Ablation study results on the THG datasets. Bold indicates the best method.}
 \vspace{-2pt}
 \resizebox{0.8\columnwidth}{!}{
 \begin{tabular}{lccccc}\toprule
    \textbf{Model} &  ROUGE-1 &  ROUGE-2 & ROUGE-L & F1@1 & F1@5 \\ \midrule
    RIGHT & \textbf{62.11} &\textbf{43.86} & \textbf{61.39} & \textbf{30.58} & \textbf{43.23} \\
    \hdashline
    ~w/o Retriever & 59.91 & 41.63 & 59.23 & 29.66 & 41.70 \\
    ~w/o Selector & 60.49 & 42.06 & 59.76 & 30.22 & 41.95\\
    ~w/o Generator & 36.24 & 16.02 & 32.86 & 24.61 & 26.73
    \\\bottomrule
 \end{tabular}}
 \vspace{-5pt}
 \label{tab:ab_study}
\end{table}

\subsubsection{Impact of the Number of Augmented Hashtags.}
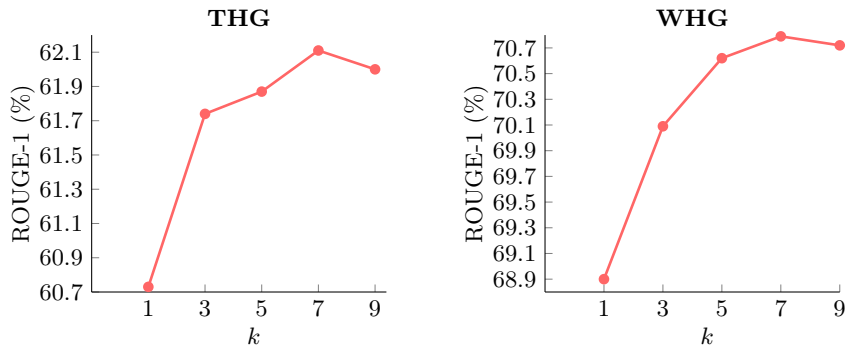
\begin{figure}[t]
    \begin{centering}
    \vspace{-2pt}
    \begin{tikzpicture}
        \pgfplotsset{footnotesize,samples=10}
        \begin{groupplot}[
            group style = {group size = 2 by 1, horizontal sep = 60pt},
            width = 5.5cm, 
            height = 5cm]
            \nextgroupplot[
                align = center,
                title = {\textbf{THG}},
                xmin=0, xmax=52,
                ymin=60.7, ymax=62.2,
                xtick={10, 20, 30, 40, 50},
                axis x line*=bottom,
                axis y line*=left,
                xticklabels={1, 3, 5, 7, 9},
                ylabel={ROUGE-1 (\%)},
                xlabel={$k$},
                ytick={60.7, 60.9, 61.1, 61.3, 61.5, 61.7, 61.9, 62.1},
                grid style=dashed,
                x label style={at={(axis description cs:0.55,0.05)},anchor=north},
                y label style={at={(axis description cs:0.15,0.5)},anchor=south},
                xtick pos=bottom,
                ytick pos=left,
                ]
                \addplot[ 
                    color=color2,
                    mark=*,
                    mark size=1.5pt,
                    line width=1pt,
                    ]
                    coordinates {
                    (10, 60.73)
                    (20, 61.74)
                    (30, 61.87)
                    (40, 62.11)
                    (50, 62.00)
                    };
            \nextgroupplot[
                align = center,
                title = {\textbf{WHG}},
                legend style={at={(-0.12,1.4)},anchor=south},
                xmin=0, xmax=50,
                ymin=68.8, ymax=70.8,
                xtick={10, 20, 30, 40, 50},
                axis x line*=bottom,
                axis y line*=left,
                xticklabels={1, 3, 5, 7, 9},
                ylabel={ROUGE-1 (\%)},
                xlabel={$k$},
                ytick={68.9, 69.1, 69.3, 69.5, 69.7, 69.9, 70.1, 70.3, 70.5, 70.7, 70.9},
                grid style=dashed,
                x label style={at={(axis description cs:0.55,0.05)},anchor=north},
                y label style={at={(axis description cs:0.15,0.5)},anchor=south},
                xtick pos=bottom,
                ytick pos=left,
                legend style={draw=none},
                legend cell align=left,
                legend style={at={(-0.29,-0.3)},anchor=south,font=\footnotesize},
                legend columns=3,
                ]
                \addplot[ 
                    color=color2,
                    mark=*,
                    mark size=1.5pt,
                    line width=1pt,
                    ]
                    coordinates {
                    (10, 68.90)
                    (20, 70.09)
                    (30, 70.62)
                    (40, 70.79)
                    (50, 70.72)
                    };
        \end{groupplot}
    \end{tikzpicture}
    \vspace{-12pt}
    \caption{Our Rouge-1 results in the different number of augmented hashtags (i.e., $k=1, 3, 5, 7, 9$).}
    \label{fig:the_number_of_k}
    \end{centering}
\end{figure}

To explore the impact of the number of concatenated hashtags with the input tweet, we conduct a series of experiments. Specifically, we concatenate various top-$k$ ($k=1, 3, 5, 7, 9$) with the input tweet for the THG and the WHG datasets. Figure~\ref{fig:the_number_of_k} demonstrates the Rouge-1 results (other metrics show the same trends), showing that:
\begin{enumerate}
    \item Retrieval augmentation aids in improving the performance when a sufficient number is considered, suggesting that augmenting more hashtags increases the probability of covering mainstream hashtags and makes the generator more robust in the presence of mismatches from certain hashtags.
    \item Upon reaching a certain threshold of the number of augmented hashtags (i.e., $k=7$), the performance converges, suggesting that the majority of mainstream hashtags might have already been augmented.
\end{enumerate}

\subsubsection{Case Study.}
To validate the successful recall of mainstream hashtags and the potential for further improvement, we analyze the successful and unsuccessful cases and present a representative case study in Table~\ref{tab:case_study}. We conclude that: 
\begin{enumerate}
    \item Some retrieved tweets share the same topic as the input tweets but have subpar labeled hashtags (e.g., ``fx logix'' in Table~\ref{tab:case_study}). Fortunately, our generator demonstrates the capability to disregard these irrelevant hashtags.
    \item Some retrieved tweets are partially relevant to the input tweet. Although the retrieved hashtags align well with the topic of the retrieved tweet, it is not highly pertinent to the primary topic of the input tweet (e.g., ``azure'' in the retrieved hashtags). Nonetheless, our generator can filter out these irrelevant hashtags.
    \item The generation model produces a semantically accurate but non-mainstream hashtag ``windows virtual desktop'' by directly copying the original word from the input tweet due to its limited knowledge of mainstream hashtags. However, our retriever and selector effectively identify the corresponding mainstream hashtag in its abbreviated form ``wvd''. Our RIGHT successfully replaced the original hashtag with the mainstream hashtag, indicating the effectiveness of retrieval augmentation.
    \item Seq2Seq generates certain hashtags that are also retrieved by the retriever, while RIGHT fails to generate them (e.g., ``microsoft''). These cases constitute less than 1\% of the total. We speculate that this discrepancy may be due to the selector placing the correct hashtags toward the end of the list, leading to reduced confidence and subsequent non-adoption by the generator.
\end{enumerate}

\begin{table}[t]
 \centering
    \vspace{-2pt}
 \caption{An example from the THG test set. Correct results are marked bold.}
 \vspace{-2pt}
 \resizebox{1\columnwidth}{!}{
 \begin{tabular}{l m{8cm}<{\centering}}\toprule
    \textbf{Input:} & Geeks guide to  microsoft teams optimization with windows virtual desktop citrix. \\ \midrule
    \textbf{Label:} & \textbf{microsoft}; \textbf{windows}; \textbf{citrix}; {\textbf{wvd}}    \\ \midrule
    \textbf{Retriever \& Selector:} & \textbf{citrix}; {\textbf{wvd}}; fs logix; v mware; azure; aws; \textbf{microsoft}    \\  \midrule
    \textbf{Seq2Seq:} & {\textbf{microsoft}}; windows virtual desktop; \textbf{citrix}; vdi   \\ \midrule
    \textbf{RIGHT:} & \textbf{citrix}; {\textbf{wvd}}
    \\\bottomrule
 \end{tabular}}
 \label{tab:case_study}
\end{table}

\vspace{-8pt}
\section{Related Works}
\label{relatedworks}
Our work mainly builds on two streams of previous work: hashtag recommendation and retrieval-augmented generation.
\subsection{Mainstream hashtag recommendation}
Mainstream hashtag recommendation aims to provide users with short topical and popular tags representing the main ideas of their tweets before publication. Three primary methods have been proposed for this task~\cite{kwak2010twitter}:

1) \textbf{Keyphrase extraction method} formulates this task as keyphrases extraction from source posts~\cite{DBLP:conf/emnlp/GongZH15,zhang-2016-keyphrase,zhang-2018-encoding}, which fails to produce hashtags that do not appear in the microblog posts while large freedom is allowed for users to write whatever hashtags they like. The performance of this method is much lower than other methods.
2) \textbf{Retrieval-based method} aims to retrieve from a predefined hashtag list \cite{weston-2014-tagspace,DBLP:conf/ijcai/GongZ16,huang-2016-hashtag,DBLP:conf/ijcai/ZhangWHHG17}, which is limited to generating only the hashtags that are included in the list. In reality, a wide range of hashtags can be created every day, resulting impossibility to be covered by a fixed list and the difficulty to maintain the list.
3) \textbf{Generation-based method} was proposed~\cite{wang-2019-microblog,DBLP:journals/corr/abs-2104-08723,MAO2022109581,ni2023comparative} to overcome the aforementioned challenges, which formulates the task as a sequence-to-sequence generation paradigm, allowing for the creation of a wider range of hashtags that better capture the main ideas of the microblog post. However, previous studies pay limited attention to mainstream
hashtags. Consequently, even though it produces semantically correct tags, the tweet might fail to be indexed by a mainstream hashtag on microblog services due to tags' unpopularity, thus weakening the recall rate of microblog searches.

To the best of our knowledge, we are the first to alleviate this issue by combining retrieval and generation methods. Meanwhile, we improve the quality of the retriever at a low cost.

\subsection{Retrieval-augmented Generation}
The retrieval-augmented generation represents a novel paradigm that merges pre-trained generative models with information retrieval techniques~\cite{asai-etal-2023-retrieval,DBLP:conf/nips/LewisPPPKGKLYR020}.
Previous research in this field primarily has focused on introducing external knowledge to address knowledge-intensive tasks~\cite{chen-2017-reading,zhang-etal-2023-relevance,he-etal-2023-merging,li2023generative,wang2022micro,wang2021discover,he2023mera,ugr,corpusbrain,clever} and utilizing similar data to enhance the model performance across various natural language processing (NLP) tasks, including image captioning~\cite{ramos-2023-retrieval}, keyphrase generation~\cite{kim-2021-structure,gao-2022-retrieval}, named entity recognition~\cite{zhang-2022-domain,gere}, and others. Recently, this technique has also been used in large language models to alleviate issues like factual hallucination~\cite{cao-etal-2020-factual,raunak-etal-2021-curious,hallucination-ji-2023}, knowledge out-dating~\cite{he2022rethinking}, and the lack of domain-specific expertise~\cite{li2023chatgpt}.

Notably, we adopt this framework to the mainstream hashtag recommendation task, by introducing a selector combining global signals to improve mainstream identification.

\section{Conclusion}
\label{sec:conclusion}
In this study, we have proposed a simple yet effective retrieval-augmented generative recommender, designed to utilize the advantage of retrieval and generation methods for mainstream hashtag recommendation. To improve the quality of the retriever's results at a low cost, we have integrated a selector module into the conventional retrieval-augmented framework. Specifically, the retriever's role is to find relevant hashtags on the same topic, the selector is employed to enhance the identification of mainstream hashtags, and the generator is responsible for combining input tweets and selected hashtags to generate desired hashtags. We have conducted extension experiments using two extensive datasets to validate the effectiveness of our approach.

In future work, it is valuable to explore optimal strategies for combining the retrieval-based method with the generation-based method, as well as developing a co-training approach that jointly refines the three components.

\section*{Acknowledgements}
This work was funded by the National Natural Science Foundation of China (NSFC) under Grants No. 62372431, and 62006218, the Youth Innovation Promotion Association CAS under Grants No. 2021100, the project under Grants No. 2023YFA1011602,  JCKY2022130C039 and 2021QY1701, and the Lenovo-CAS Joint Lab Youth Scientist Project. All content represents the opinion of the authors, which is not necessarily shared or endorsed by their respective employers and/or sponsors.

%
%
%

\bibliographystyle{splncs04}
\bibliography{custom}

\begin{thebibliography}{10}
\providecommand{\url}[1]{\texttt{#1}}
\providecommand{\urlprefix}{URL }
\providecommand{\doi}[1]{https://doi.org/#1}

\bibitem{asai-etal-2023-retrieval}
Asai, A., Min, S., Zhong, Z., Chen, D.: Retrieval-based language models and applications. In: ACL. pp. 41--46. Toronto, Canada (Jul 2023). \doi{10.18653/v1/2023.acl-tutorials.6}, \url{https://aclanthology.org/2023.acl-tutorials.6}

\bibitem{DBLP:conf/nips/BrownMRSKDNSSAA20}
Brown, T.B., Mann, B., Ryder, N., Subbiah, M., Kaplan, J., Dhariwal, P., Neelakantan, A., Shyam, P., Sastry, G., Askell, A., Agarwal, S., Herbert{-}Voss, A., Krueger, G., Henighan, T., Child, R., Ramesh, A., Ziegler, D.M., Wu, J., Winter, C., Hesse, C., Chen, M., Sigler, E., Litwin, M., Gray, S., Chess, B., Clark, J., Berner, C., McCandlish, S., Radford, A., Sutskever, I., Amodei, D.: Language models are few-shot learners. In: Larochelle, H., Ranzato, M., Hadsell, R., Balcan, M., Lin, H. (eds.) NeurIPS (2020), \url{https://proceedings.neurips.cc/paper/2020/hash/1457c0d6bfcb4967418bfb8ac142f64a-Abstract.html}

\bibitem{cao-etal-2020-factual}
Cao, M., Dong, Y., Wu, J., Cheung, J.C.K.: Factual error correction for abstractive summarization models. In: EMNLP (Nov 2020). \doi{10.18653/v1/2020.emnlp-main.506}, \url{https://aclanthology.org/2020.emnlp-main.506}

\bibitem{chen-2017-reading}
Chen, D., Fisch, A., Weston, J., Bordes, A.: Reading {W}ikipedia to answer open-domain questions. In: ACL (Jul 2017). \doi{10.18653/v1/P17-1171}, \url{https://aclanthology.org/P17-1171}

\bibitem{gere}
Chen, J., Zhang, R., Guo, J., Fan, Y., Cheng, X.: {GERE:} generative evidence retrieval for fact verification. In: {SIGIR} 2022. pp. 2184--2189. {ACM} (2022). \doi{10.1145/3477495.3531827}, \url{https://doi.org/10.1145/3477495.3531827}

\bibitem{corpusbrain}
Chen, J., Zhang, R., Guo, J., Liu, Y., Fan, Y., Cheng, X.: Corpusbrain: Pre-train a generative retrieval model for knowledge-intensive language tasks. In: CIKM 2022. pp. 191--200. {ACM} (2022). \doi{10.1145/3511808.3557271}, \url{https://doi.org/10.1145/3511808.3557271}

\bibitem{clever}
Chen, J., Zhang, R., Guo, J., de~Rijke, M., Chen, W., Fan, Y., Cheng, X.: Continual learning for generative retrieval over dynamic corpora. In: CIKM 2023. pp. 306--315. {ACM} (2023). \doi{10.1145/3583780.3614821}, \url{https://doi.org/10.1145/3583780.3614821}

\bibitem{ugr}
Chen, J., Zhang, R., Guo, J., de~Rijke, M., Liu, Y., Fan, Y., Cheng, X.: A unified generative retriever for knowledge-intensive language tasks via prompt learning. In: SIGIR 2023. pp. 1448--1457. {ACM} (2023). \doi{10.1145/3539618.3591631}, \url{https://doi.org/10.1145/3539618.3591631}

\bibitem{devlin-2019-bert}
Devlin, J., Chang, M.W., Lee, K., Toutanova, K.: {BERT}: Pre-training of deep bidirectional transformers for language understanding. In: NAACL (Jun 2019). \doi{10.18653/v1/N19-1423}, \url{https://aclanthology.org/N19-1423}

\bibitem{ding-2012-automatic}
Ding, Z., Zhang, Q., Huang, X.: Automatic hashtag recommendation for microblogs using topic-specific translation model. In: COLING (Dec 2012), \url{https://aclanthology.org/C12-2027}

\bibitem{gao-2021-simcse}
Gao, T., Yao, X., Chen, D.: {S}im{CSE}: Simple contrastive learning of sentence embeddings. In: EMNLP (Nov 2021). \doi{10.18653/v1/2021.emnlp-main.552}, \url{https://aclanthology.org/2021.emnlp-main.552}

\bibitem{gao-2022-retrieval}
Gao, Y., Yin, Q., Li, Z., Meng, R., Zhao, T., Yin, B., King, I., Lyu, M.: Retrieval-augmented multilingual keyphrase generation with retriever-generator iterative training. In: NAACL (2022). \doi{10.18653/v1/2022.findings-naacl.92}, \url{https://aclanthology.org/2022.findings-naacl.92}

\bibitem{gong-2015-hashtag}
Gong, Y., Zhang, Q., Huang, X.: Hashtag recommendation using {D}irichlet process mixture models incorporating types of hashtags. In: EMNLP (2015). \doi{10.18653/v1/D15-1046}, \url{https://aclanthology.org/D15-1046}

\bibitem{DBLP:conf/emnlp/GongZH15}
Gong, Y., Zhang, Q., Huang, X.: Hashtag recommendation using dirichlet process mixture models incorporating types of hashtags. In: M{\`{a}}rquez, L., Callison{-}Burch, C., Su, J., Pighin, D., Marton, Y. (eds.) EMNLP (2015). \doi{10.18653/v1/d15-1046}, \url{https://doi.org/10.18653/v1/d15-1046}

\bibitem{DBLP:conf/ijcai/GongZ16}
Gong, Y., Zhang, Q.: Hashtag recommendation using attention-based convolutional neural network. In: Kambhampati, S. (ed.) IJCAI (2016), \url{http://www.ijcai.org/Abstract/16/395}

\bibitem{he2022rethinking}
He, H., Zhang, H., Roth, D.: Rethinking with retrieval: Faithful large language model inference. arXiv preprint  (2022), \url{https://arxiv.org/pdf/2301.00303.pdf}

\bibitem{he2023mera}
He, S., Fan, R.Z., Ding, L., Shen, L., Zhou, T., Tao, D.: Mera: Merging pretrained adapters for few-shot learning. arXiv preprint arXiv:2308.15982  (2023), \url{https://arxiv.org/abs/2308.15982}

\bibitem{he-etal-2023-merging}
He, S., Fan, R.Z., Ding, L., Shen, L., Zhou, T., Tao, D.: Merging experts into one: Improving computational efficiency of mixture of experts. In: Bouamor, H., Pino, J., Bali, K. (eds.) Proceedings of the 2023 Conference on Empirical Methods in Natural Language Processing. pp. 14685--14691. Association for Computational Linguistics, Singapore (Dec 2023), \url{https://aclanthology.org/2023.emnlp-main.907}

\bibitem{huang-2016-hashtag}
Huang, H., Zhang, Q., Gong, Y., Huang, X.: Hashtag recommendation using end-to-end memory networks with hierarchical attention. In: COLING (Dec 2016), \url{https://aclanthology.org/C16-1090}

\bibitem{hallucination-ji-2023}
Ji, Z., Lee, N., Frieske, R., Yu, T., Su, D., Xu, Y., Ishii, E., Bang, Y.J., Madotto, A., Fung, P.: Survey of hallucination in natural language generation. ACM Comput. Surv.  (2023). \doi{10.1145/3571730}, \url{https://doi.org/10.1145/3571730}

\bibitem{kim-2021-structure}
Kim, J., Jeong, M., Choi, S., Hwang, S.w.: Structure-augmented keyphrase generation. In: EMNLP (2021). \doi{10.18653/v1/2021.emnlp-main.209}, \url{https://aclanthology.org/2021.emnlp-main.209}

\bibitem{DBLP:journals/corr/KingmaB14}
Kingma, D.P., Ba, J.: Adam: {A} method for stochastic optimization. In: Bengio, Y., LeCun, Y. (eds.) ICLR (2015), \url{http://arxiv.org/abs/1412.6980}

\bibitem{ko2023chatgpt}
Kocoń, J., Cichecki, I., Kaszyca, O., Kochanek, M., Szydło, D., Baran, J., Bielaniewicz, J., Gruza, M., Janz, A., Kanclerz, K., Kocoń, A., Koptyra, B., Mieleszczenko-Kowszewicz, W., Miłkowski, P., Oleksy, M., Piasecki, M., Łukasz Radliński, Wojtasik, K., Woźniak, S., Kazienko, P.: Chatgpt: Jack of all trades, master of none. arXiv preprint  (2023), \url{https://arxiv.org/pdf/2302.10724.pdf}

\bibitem{kwak2010twitter}
Kwak, H., Lee, C., Park, H., Moon, S.: What is twitter, a social network or a news media? In: WWW (2010), \url{https://ink.library.smu.edu.sg/cgi/viewcontent.cgi?article=7104&context=sis_research}

\bibitem{DBLP:conf/acl/LewisLGGMLSZ20}
Lewis, M., Liu, Y., Goyal, N., Ghazvininejad, M., Mohamed, A., Levy, O., Stoyanov, V., Zettlemoyer, L.: {BART:} denoising sequence-to-sequence pre-training for natural language generation, translation, and comprehension. In: Jurafsky, D., Chai, J., Schluter, N., Tetreault, J.R. (eds.) ACL (2020). \doi{10.18653/v1/2020.acl-main.703}, \url{https://doi.org/10.18653/v1/2020.acl-main.703}

\bibitem{DBLP:conf/nips/LewisPPPKGKLYR020}
Lewis, P.S.H., Perez, E., Piktus, A., Petroni, F., Karpukhin, V., Goyal, N., K{\"{u}}ttler, H., Lewis, M., Yih, W., Rockt{\"{a}}schel, T., Riedel, S., Kiela, D.: Retrieval-augmented generation for knowledge-intensive {NLP} tasks. In: Larochelle, H., Ranzato, M., Hadsell, R., Balcan, M., Lin, H. (eds.) NeurIPS (2020), \url{https://proceedings.neurips.cc/paper/2020/hash/6b493230205f780e1bc26945df7481e5-Abstract.html}

\bibitem{li2023generative}
Li, J., Sun, S., Yuan, W., Fan, R.Z., Zhao, H., Liu, P.: Generative judge for evaluating alignment. arXiv preprint arXiv:2310.05470  (2023), \url{https://arxiv.org/abs/2310.05470}

\bibitem{li2023chatgpt}
Li, X., Zhu, X., Ma, Z., Liu, X., Shah, S.: Are chatgpt and gpt-4 general-purpose solvers for financial text analytics? an examination on several typical tasks. arXiv preprint  (2023), \url{https://arxiv.org/pdf/2305.05862.pdf}

\bibitem{liu2019roberta}
Liu, Y., Ott, M., Goyal, N., Du, J., Joshi, M., Chen, D., Levy, O., Lewis, M., Zettlemoyer, L., Stoyanov, V.: Roberta: A robustly optimized bert pretraining approach. arXiv preprint  (2019), \url{https://arxiv.org/pdf/1907.11692.pdf}

\bibitem{MAO2022109581}
Mao, Q., Li, X., Liu, B., Guo, S., Hao, P., Li, J., Wang, L.: Attend and select: A segment selective transformer for microblog hashtag generation. Knowledge-Based Systems  (2022). \doi{https://doi.org/10.1016/j.knosys.2022.109581}, \url{https://www.sciencedirect.com/science/article/pii/S0950705122007973}

\bibitem{mialon2023augmented}
Mialon, G., Dess{\`\i}, R., Lomeli, M., Nalmpantis, C., Pasunuru, R., Raileanu, R., Rozi{\`e}re, B., Schick, T., Dwivedi-Yu, J., Celikyilmaz, A., et~al.: Augmented language models: a survey. arXiv preprint  (2023), \url{https://arxiv.org/pdf/2302.07842.pdf}

\bibitem{ni2023comparative}
Ni, S., Bi, K., Guo, J., Cheng, X.: A comparative study of training objectives for clarification facet generation. In: Annual International ACM SIGIR Conference on Research and Development in Information Retrieval in the Asia Pacific Region. pp. 1--10 (2023)

\bibitem{radford2018improving}
Radford, A., Narasimhan, K., Salimans, T., Sutskever, I., et~al.: Improving language understanding by generative pre-training. preprint  (2018), \url{https://cdn.openai.com/research-covers/language-unsupervised/language_understanding_paper.pdf}

\bibitem{radford2019language}
Radford, A., Wu, J., Child, R., Luan, D., Amodei, D., Sutskever, I., et~al.: Language models are unsupervised multitask learners. preprint  (2019), \url{https://cdn.openai.com/better-language-models/language_models_are_unsupervised_multitask_learners.pdf}

\bibitem{DBLP:journals/jmlr/RaffelSRLNMZLL20}
Raffel, C., Shazeer, N., Roberts, A., Lee, K., Narang, S., Matena, M., Zhou, Y., Li, W., Liu, P.J.: Exploring the limits of transfer learning with a unified text-to-text transformer. J. Mach. Learn. Res.  (2020), \url{http://jmlr.org/papers/v21/20-074.html}

\bibitem{raunak-etal-2021-curious}
Raunak, V., Menezes, A., Junczys-Dowmunt, M.: The curious case of hallucinations in neural machine translation. In: ACL (Jun 2021). \doi{10.18653/v1/2021.naacl-main.92}, \url{https://aclanthology.org/2021.naacl-main.92}

\bibitem{ramos-2023-retrieval}
Rita~Ramos, Desmond~Elliott, B.M.: Retrieval-augmented image captioning. In: EACL (2023), \url{https://arxiv.org/abs/2302.08268}

\bibitem{DBLP:conf/sigir/RobertsonW94}
Robertson, S.E., Walker, S.: Some simple effective approximations to the 2-poisson model for probabilistic weighted retrieval. In: Croft, W.B., van Rijsbergen, C.J. (eds.) SIGIR (1994). \doi{10.1007/978-1-4471-2099-5\_24}, \url{https://doi.org/10.1007/978-1-4471-2099-5\_24}

\bibitem{wang2022micro}
Wang, S., Gan, T., Liu, Y., Wu, J., Cheng, Y., Nie, L.: Micro-influencer recommendation by multi-perspective account representation learning. IEEE Transactions on Multimedia  (2022)

\bibitem{wang2021discover}
Wang, S., Gan, T., Liu, Y., Zhang, L., Wu, J., Nie, L.: Discover micro-influencers for brands via better understanding. IEEE Transactions on Multimedia  \textbf{24},  2595--2605 (2021)

\bibitem{wang-2022-training}
Wang, S., Xu, Y., Fang, Y., Liu, Y., Sun, S., Xu, R., Zhu, C., Zeng, M.: Training data is more valuable than you think: A simple and effective method by retrieving from training data. In: ACL (May 2022). \doi{10.18653/v1/2022.acl-long.226}, \url{https://aclanthology.org/2022.acl-long.226}

\bibitem{wang-2019-microblog}
Wang, Y., Li, J., King, I., Lyu, M.R., Shi, S.: Microblog hashtag generation via encoding conversation contexts. In: NAACL (Jun 2019). \doi{10.18653/v1/N19-1164}, \url{https://aclanthology.org/N19-1164}

\bibitem{weston-2014-tagspace}
Weston, J., Chopra, S., Adams, K.: {\#}{T}ag{S}pace: Semantic embeddings from hashtags. In: EMNLP (Oct 2014). \doi{10.3115/v1/D14-1194}, \url{https://aclanthology.org/D14-1194}

\bibitem{DBLP:conf/naacl/XueCRKASBR21}
Xue, L., Constant, N., Roberts, A., Kale, M., Al{-}Rfou, R., Siddhant, A., Barua, A., Raffel, C.: mt5: {A} massively multilingual pre-trained text-to-text transformer. In: Toutanova, K., Rumshisky, A., Zettlemoyer, L., Hakkani{-}T{\"{u}}r, D., Beltagy, I., Bethard, S., Cotterell, R., Chakraborty, T., Zhou, Y. (eds.) NAACL (2021). \doi{10.18653/v1/2021.naacl-main.41}, \url{https://doi.org/10.18653/v1/2021.naacl-main.41}

\bibitem{zhang-etal-2023-relevance}
Zhang, H., Zhang, R., Guo, J., de~Rijke, M., Fan, Y., Cheng, X.: From relevance to utility: Evidence retrieval with feedback for fact verification. In: Bouamor, H., Pino, J., Bali, K. (eds.) Findings of the Association for Computational Linguistics: EMNLP 2023. pp. 6373--6384. Association for Computational Linguistics, Singapore (Dec 2023), \url{https://aclanthology.org/2023.findings-emnlp.422}

\bibitem{DBLP:conf/ijcai/ZhangWHHG17}
Zhang, Q., Wang, J., Huang, H., Huang, X., Gong, Y.: Hashtag recommendation for multimodal microblog using co-attention network. In: Sierra, C. (ed.) IJCAI (2017). \doi{10.24963/ijcai.2017/478}, \url{https://doi.org/10.24963/ijcai.2017/478}

\bibitem{zhang-2016-keyphrase}
Zhang, Q., Wang, Y., Gong, Y., Huang, X.: Keyphrase extraction using deep recurrent neural networks on {T}witter. In: EMNLP (Nov 2016). \doi{10.18653/v1/D16-1080}, \url{https://aclanthology.org/D16-1080}

\bibitem{zhang-2022-domain}
Zhang, X., Jiang, Y., Wang, X., Hu, X., Sun, Y., Xie, P., Zhang, M.: Domain-specific {NER} via retrieving correlated samples. In: COLING (2022), \url{https://aclanthology.org/2022.coling-1.211}

\bibitem{zhang-2018-encoding}
Zhang, Y., Li, J., Song, Y., Zhang, C.: Encoding conversation context for neural keyphrase extraction from microblog posts. In: NAACL (Jun 2018). \doi{10.18653/v1/N18-1151}, \url{https://aclanthology.org/N18-1151}

\bibitem{DBLP:journals/corr/abs-2104-08723}
Zheng, X., Mekala, D., Gupta, A., Shang, J.: News meets microblog: Hashtag annotation via retriever-generator. arXiv preprint  (2021), \url{https://arxiv.org/abs/2104.08723}

\end{thebibliography}
%




\end{document}